\documentclass[letterpaper, 10 pt, conference]{ieeeconf}  

\IEEEoverridecommandlockouts                              

\usepackage{amssymb}  
\usepackage{bm}
\usepackage{multirow} 
\usepackage{graphicx} 
\usepackage{svg} 
\usepackage{amsmath} 
\usepackage{amsfonts} 
\def\mathbi#1{\textbf{\em #1}} 
\usepackage{booktabs} 
\usepackage{cite} 
\usepackage{amsmath}
\usepackage{pdfpages} 
\usepackage{adjustbox}
\usepackage[font=footnotesize]{caption} 

\title{\LARGE \bf
CrossDTR: Cross-view and Depth-guided Transformers\\for 3D Object Detection
}

\author{Ching-Yu Tseng$^{1}$, Yi-Rong Chen$^{1}$, Hsin-Ying Lee$^{1}$,  Tsung-Han Wu$^{1}$, Wen-Chin Chen$^{1}$, and Winston H. Hsu$^{1, 2}$
\thanks{$^{1}$National Taiwan University, $^{2}$Mobile Drive Technology}
}

\begin{document}

\maketitle
\thispagestyle{empty}
\pagestyle{empty}


\begin{abstract}
To achieve accurate 3D object detection at a low cost for autonomous driving, many multi-camera methods have been proposed and solved the occlusion problem of monocular approaches. However, due to the lack of accurate estimated depth, existing multi-camera methods often generate multiple bounding boxes along a ray of depth direction for difficult small objects such as pedestrians, resulting in an extremely low recall. Furthermore, directly applying depth prediction modules to existing multi-camera methods, generally composed of large network architectures, cannot meet the real-time requirements of self-driving applications. To address these issues, we propose Cross-view and Depth-guided Transformers for 3D Object Detection, CrossDTR. First, our lightweight \textit{depth predictor} is designed to produce precise object-wise sparse depth maps and low-dimensional depth embeddings without extra depth datasets during supervision. Second, a \textit{cross-view depth-guided transformer} is developed to fuse the depth embeddings as well as image features from cameras of different views and generate 3D bounding boxes. Extensive experiments demonstrated that our method hugely surpassed existing multi-camera methods by 10 percent in pedestrian detection and about 3 percent in overall mAP and NDS metrics. Also, computational analyses showed that our method is 5 times faster than prior approaches. Our codes will be made publicly available at https://github.com/sty61010/CrossDTR.
\end{abstract}

\section{Introduction}
Detecting instances of objects in the 3D space from sensor information, i.e. 3D object detection, is crucial for various intelligent systems, such as autonomous driving and indoor robotics. Previous work tends to rely on accurate depth information from different sensors, such as LiDAR signals \cite{second, pointpillars, centepoint} and binocular information \cite{dsgn, stereorcnn}, to accomplish superior performance.
In recent years, in order to achieve high-quality detection at a low cost, several methods based on commodity cameras have been proposed. Among them, since naive monocular detection \cite{oft, centernet, pseudolidar, pseudolidarv2, caddn, fcos3d, pgd} suffers from the problems of occlusion and deficiency of cross-view information, methods transforming camera information from multiple views into Bird-Eye-View \cite{detr3d, bevdet, bevdet4d, bevformer, petr, petrv2, graphdetr3d, beverse, imvoxelnet}, called multi-view methods, has received increasing attention.

\begin{figure}[t]
    \centering
    \includegraphics[width=0.45\textwidth, trim=0 0 0 0, clip]{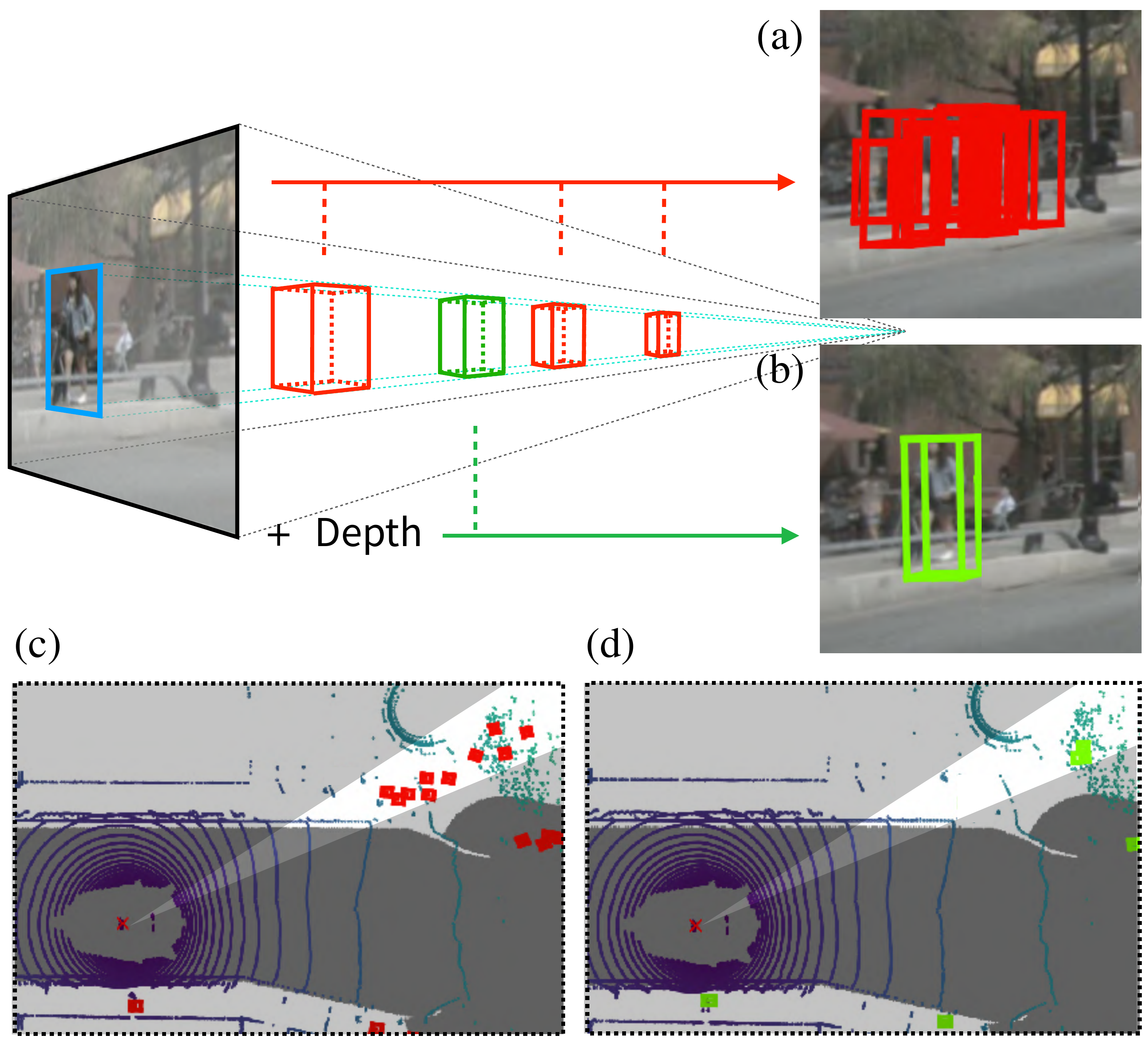}
    \caption{{\bf Multi-camera methods suffer from inaccurate depth estimation.} Red and green bounding boxes represent inaccurate and accurate predictions respectively. The above 2D-to-3D projection diagram mainly shows that (a) previous multi-view methods usually produce a row of false positives predictions alone a ray of depth, but (b) our method, guided with depth hints, can precisely predict only one bounding box. Plot (c) and (d) demonstrate the corresponding bird-eye view predictions of (a) and (b).}
    \label{fig: overview}
\end{figure}

Though these multi-view methods have made some progress with cross-view information and Bird-Eye-View representation \cite{lss, fiery, bevdet, bevdet4d, bevformer, cvt}, we observe existing practices suffered from either extremely low recall in small objects due to inaccurate depth or an unaffordable computational burden because of complex depth prediction modules. Specifically, while some methods fusing information from multiple views \cite{detr3d, bevdet, bevdet4d, bevformer, petr, petrv2, graphdetr3d, beverse, imvoxelnet} easily locate the pixel coordinate of small objects in images, they can hardly estimate precise distances of objects from the image plane. Consequently, these methods tend to predict a row of false positive bounding boxes along a ray of depth direction in candidate regions when detecting small objects (shown in Fig. \ref{fig: overview}), leading to low recall in perception and poor follow-up prediction and planning. In addition, some previous monocular approaches utilized complex depth prediction modules \cite{monogeo, d4lcn} or large-scale depth-pretrained backbone \cite{ddad, dd3d, vovnet, vovnetv2} to provide depth cues. Nevertheless, directly applying them to existing multi-camera methods, generally composed of large network architectures, cannot meet the real-time requirements of the self-driving application (Tab. \ref{tab: sota multi-cam}). From the two observations above, we conclude that a module is needed to obtain depth hints from multiple cameras and fuse both depth and image information from different views in real time.

\begin{figure*}[!t]
    \centering
    \includegraphics[width=\textwidth]{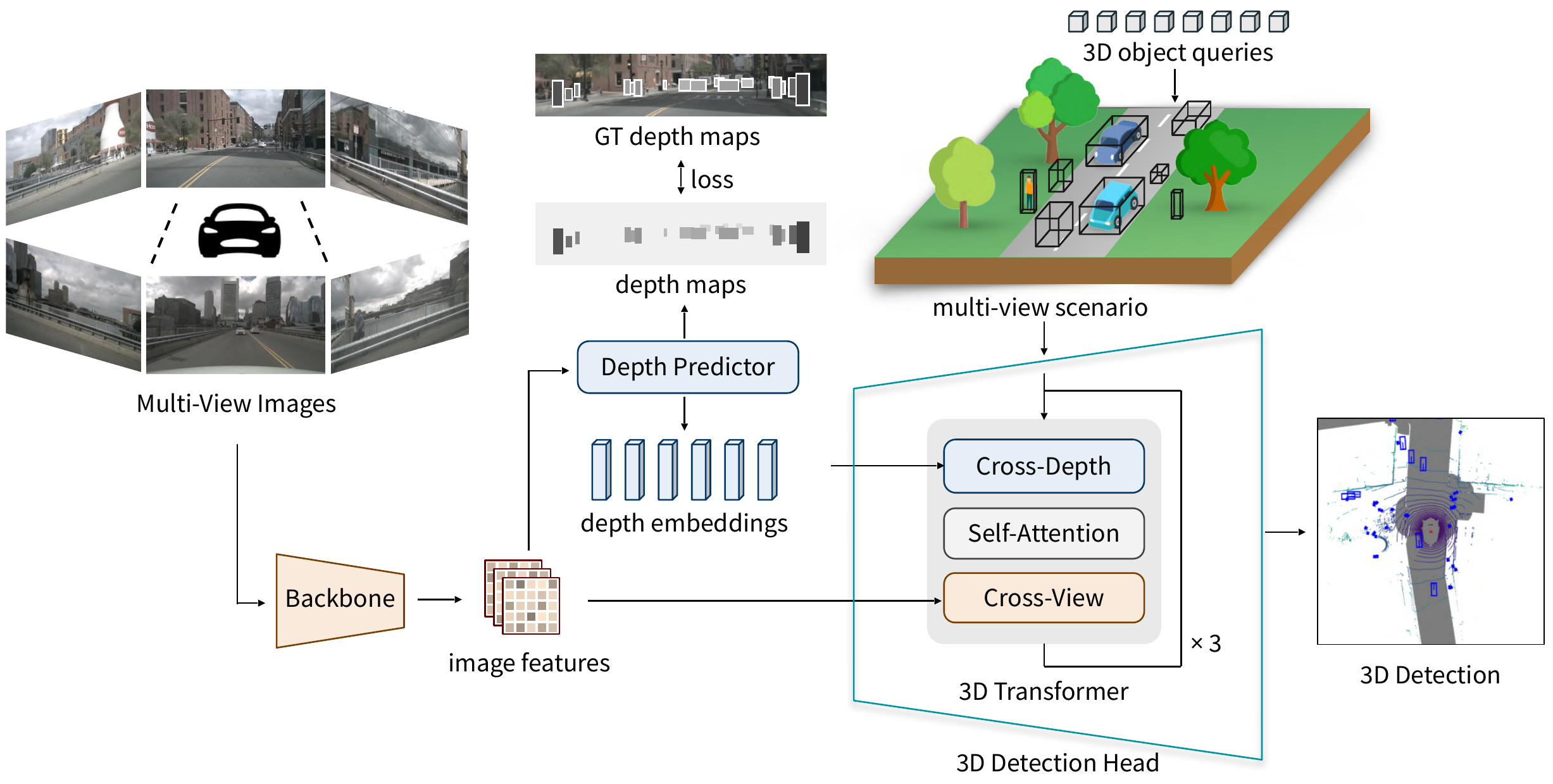}

    \caption{{\bf The overall framework of our proposed CrossDTR.} First, Multi-view images are fed into a feature extractor backbone to generate image features. Then, image features are fed into our Depth Predictor (in Sec. \ref{depth predictor}) to produce depth embeddings and predicted depth maps. Lastly, given image features, depth embeddings, and 3D object queries, our Cross-view and Depth-guided Transformer (in Sec. \ref{depth-guided transformer}) conducts cross-view attention and cross-depth attention to generate 3D bounding boxes. Note that we minimize the difference between predicted depths and our generated object-wise sparse depth maps (in Sec. \ref{object-wise sparse depth map}) under supervision during training. Best viewed in color.}
    \label{fig: model architecture}
\end{figure*}

To achieve the goal, we proposed CrossDTR, a novel end-to-end Cross-view and Depth-guided Transformer network for multi-camera 3D object detection as shown in Fig. \ref{fig: model architecture}. To efficiently obtain depth hints for downstream 3D object detection, we leverage a lightweight \textit{depth predictor} to produce precise depth maps for each view (Sec. \ref{depth predictor}). Specifically, inspired by previous depth-aware methods \cite{monodetr, monodtr, monogeo, d4lcn, deviant}, the depth predictor is supervised with our generated object-wise sparse depth maps without extra depth dataset (Sec. \ref{object-wise sparse depth map}). Then, to fused the depth and image information from multi-view cameras effectively, we propose a novel \textit{cross-view and depth-guided transformer} (Sec. \ref{depth-guided transformer}). In short, the Transformer Encoder is used to compress high-resolution depth maps into low-dimensional depth embeddings, and the Transformer Decoder performs the cross-attention mechanism among depth as well as image information from multi-views.

Experimental results demonstrated that our depth-guided method resolves the problem of false positive predictions on small objects (Fig. \ref{fig: overview}) and achieves overall improvement with the limited computational burden. Compared with existing multi-camera methods on the nuScenes dataset \cite{nuscenes}, we increased by 10 percent Average precision (AP) in pedestrian detection and about 3 percent in mAP and NDS metrics. Also, computational analyses demonstrated that our lightweight method is 5 times faster than prior methods under similar network backbones. To sum up, the overall contributions of this work can be summarized as follows:

\begin{itemize}
    \item We build up a novel cross-view and depth-guided perception framework, CrossDTR, to insert accurate depth cues into multi-view detection methods.
    \item Our proposed depth-guided module can alleviate the problem of false positive predictions along the direction of depth for small objects.
    \item Our framework achieves state-of-the-art 3D detection performance on the nuScenes dataset \cite{nuscenes} with fewer computational budgets compared with existing multi-view or depth-guided methods.
\end{itemize}

\section{Related Work}
\label{sec: related work}
\subsection{Monocular 3D Object Detection}
Monocular 3D object detection \cite{oft, pseudolidar, pseudolidarv2, caddn, fcos3d, pgd} has received lots of attention due to the low cost of commodity cameras. It originates from Orthographic Feature Transform (OFT) \cite{oft}, which projects camera features into ego pose coordinate uniformly, voxelized the features, and uses the detector from PointPillar \cite{pointpillars}. OFT \cite{oft} is the first method that deals with camera features in a LiDAR-based \cite{second, pointpillars, centepoint} technique, but OFT \cite{oft} predicts the depth value of each pixel uniformly and thus results in inaccurate depth estimation. Extend from OFT, Pseudo-lidar \cite{pseudolidar, pseudolidarv2} and CaDDN \cite{caddn} methods use a convolutional neural network to predict depth distribution and auxiliary loss to enhance performance. Apart from the above methods, the other monocular methods \cite{fcos3d, pgd} directly regression 3D representation in camera coordinate. FCOS3D \cite{fcos3d} is built on FCOS \cite{fcos}, which is a single-stage 2D object detection framework. Furthermore, PGD \cite{pgd} is an improved version of FCOS3D \cite{fcos3d} with extra depth information. In conclusion, depth information is critical for monocular methods to enhance performance.

\subsection{Multi-Camera 3D Object Detection}
Multi-camera methods \cite{detr3d, bevdet, bevdet4d, imvoxelnet, bevformer, petr, petrv2, graphdetr3d, beverse} have been proved to solve the problem of occlusion through temporal and spatial information. DETR-based methods \cite{detr3d, petr, petrv2, bevformer} were first proposed. DETR3D \cite{detr3d} was built based on DETR \cite{detr} and utilized the attention mechanism to select features from different camera. Unlike DETR \cite{detr} in the 2D space, we can not assure the candidate area of objects in the 3D space, so DETR3D \cite{detr3d} initialize object query randomly with uniform distribution and pick features by projecting object position into camera coordinate. As an improved version of DETR3D \cite{detr3d}, PETR \cite{petr} enhance performance by initializing object queries with the method from Anchor DETR \cite{anchordetr} and applying 3D positional embedding to embed 3D coordinate frustum into multi-head attention. PETRv2 \cite{petrv2} optimized PETR by adding temporal information. Additionally, some researchers regard that Bird-Eye-View (BEV) \cite{vae, vpn, lss, fiery, cvt, translatingintomaps, bevsegformer} representations can provide better space concepts in 3D coordinates. BEVDet methods \cite{bevdet, bevdet4d} transform image features into BEV according to Lift-Splat-Shoot (LSS) Method \cite{lss} and propose BEV data augmentation \cite{bevdet} to avoid over-fitting. BEVFormer \cite{bevformer} proposed BEV query and use the deformable attention \cite{deformabledetr} to suppress computation. BEVFormer\cite{bevformer} also applies temporal information by cross-attention between BEV queries from different time stamps. However, those methods still lack accurate depth estimation.

\subsection{Depth-guided Monocular Methods}
Monocular methods can not achieve competitive performance in comparison with LiDAR-based methods \cite{second, pointpillars, centepoint} and binocular methods \cite{dsgn, stereorcnn}. The main reason is inaccurate depth estimation, and thus the depth-aware methods \cite{astransformer, d4lcn, monogeo, monodtr, monodetr, deviant} are proposed recently. ASTransformer \cite{astransformer} first proposed to use depth maps supervision to enhance the performance of depth estimation. Besides, both MonoDTR \cite{monodtr} and MonoDETR \cite{monodetr} compose depth-aware embedding from depth maps and are supervised by ground truth depth maps to enhance the performance of monocular object detection. However, all the above methods still build on monocular methods and will suffer from complex post-processing between cameras to aggregate all information and remove repeated predicted bounding boxes. Multi-camera method with a depth-guided module is still missing so far.

\section{Method}
\subsection{Problem Definition}
\label{problem definition}
In this work, we aim to precisely detect instances of objects in 3D space given multiple scanned RGB images \cite{survey}. Specifically, let $\mathbf{I}_{sensors} = \{\mathcal{I}_1,...,\mathcal{I}_{N_{cameras}}\}$ represent a set of scanned multi-view images, and $\mathbf{{B}}_{LiDAR} = \{\mathcal{{B}}_1,...,\mathcal{{B}}_{N_{box}}\}$ denotes a set of ground truth bounding boxes, a 3D bounding box $\mathcal{\hat{B}}_{i} \in \mathbf{{B}}_{LiDAR}$ is formulated as a vector with 7 degree of freedom:
\begin{equation}
    \mathcal{\hat{B}}_{i} = ({x}_{c}, {y}_{c}, {z}_{c}, {l}, {w}, {h}, {\theta}),
    \label{eq: bbox}
\end{equation}
where (${x}_{c}, {y}_{c}, {z}_{c}$) denotes the center of each bounding box. (${l}, {w}, {h}$) represents the length, width, and height of the cuboid respectively. ${\theta}$ means orientation (yaw) of each object. Formally, given a set of predicted bounding boxes $\mathbf{\hat{B}}_{LiDAR}$, a multi-view 3D object detector $\mathbi{f}_{Det}$ is defined as follows:
\begin{equation}
    \mathbf{\hat{B}}_{LiDAR} = \mathbi{f}_{Det}(\mathbf{I}_{sensors}).
    \label{eq: definition}
\end{equation}

\subsection{Overall Architecture}
\label{overall architecture}
Fig. \ref{fig: model architecture} illustrates our architecture. First, we feed our multi-view images $\{\mathcal{I}_1,...,\mathcal{I}_{N_{cameras}}\}$ into our model. Without external data, \cite{resnet} is applied as our backbone to extract features $\mathcal{F}_{view}$ for each view. Then, we feed image features $\mathcal{F}_{view}$ into Depth Predictor (in Sec. \ref{depth predictor}). Given single-view image features $\mathcal{F}_{view}$, the Depth Predictor produces low-dimensional depth embeddings and depth maps by a Transformer encoder. During training, we minimize the difference between the predicted depth maps and our generated sparse depth maps (in Sec. \ref{object-wise sparse depth map}) in a supervised manner. Lastly, given image features, depth embeddings, and 3D object queries, our Cross-view and Depth-guided Transformer (in Sec. \ref{depth-guided transformer}) conducts cross-view attention and cross-depth attention to generate 3D bounding boxes.

\subsection{Object-wise Sparse Depth Map}
\label{object-wise sparse depth map}
Unlike some prior depth-guided monocular methods requiring costly dense depth maps during training, we extend MonoDETR \cite{monodetr} to multi-view scenarios and leverage only the sparse depth hints provided by the raw LiDAR data, which is more cost-effective. Specifically, we first generate object-wise sparse depth maps from each view and then train our depth-guided multi-camera 3D object detector with these sparse depth hints. Although our sparse depth maps contain less information than dense maps, they are sufficient to guide a 3D detector as they provide hints of the existence of objects in certain areas with relative depth values. We first detail our depth generation process below.

Given a camera matrix $T \in \mathbb{R}^{3 \times 4}$ and a point $p \in \mathbb{R}^3$ in the LiDAR coordinate, we define the transformation function $\mathcal{T}$ from the LiDAR coordinate to the camera coordinate as follows:
\begin{equation} \label{eq: transform_function}
\begin{split}
    \mathcal{T}(T, p) &= \begin{bmatrix}
        u & v & d
    \end{bmatrix}^{T},\\
    \text{where } d \cdot \begin{bmatrix}
        u & v & 1
    \end{bmatrix}^T &= T \cdot (p \oplus 1),
\end{split}
\end{equation}
and $\oplus$ denotes tensor concatenation.

Given a set of ground truth bounding boxes $\mathbf{B}_{LiDAR}$ in the LiDAR coordinate, and the camera matrices set 
$\mathbf{T} = \{ T_m \in \mathbb{R}^{3 \times 4} \mid m=1, \ldots, N_{cameras}\}$, 
we transform the center point and corners of each bounding box into each camera view by (\ref{eq: transform_function}).
Let $p^{m,i}_{centers}, d^{m,i}_{centers}, \mathbf{P}^{m,i}_{corners}$ be the center point, the depth value of the center, and the set of corner points of the bounding box $\mathcal{B}_i$ in the $m$-th camera coordinate, then
\begin{align}
    p^{m,i}_{centers} &=  
    \begin{bmatrix}
        u_c & v_c & d_c
    \end{bmatrix}^T = 
    \mathcal{T}(T_m, p_i), \\
    d^{m,i}_{centers} &= d_c,\\
    \mathbf{P}^{m,i}_{corners} &= \{ \mathcal{T}(T_m, p) \mid p \in \mathcal{C}(\mathcal{B}_i)  \},
\end{align}
where $T_m \in \mathbf{T}$,
$p_i = \begin{bmatrix}
        x_c & y_c & z_c
       \end{bmatrix}^T$
and $x_c, y_c, z_c \in \mathcal{B}_i$.
$\mathcal{C}(\mathcal{B})$ is the function returning $8$ corners of the given 3D bounding box $\mathcal{B}$. 
We further extract the 2D bounding box $\mathcal{B}^{2d}_{m,i} = (u^i_{min}, u^i_{max}, v^i_{min}, v^i_{max})$ with respect to $\mathcal{B}_i$ in the $m$-th camera from $\mathbf{P}^{m,i}_{corners}$ by getting the minimum and maximum $(u, v)$ value in $\mathbf{P}^{m,i}_{corners}$.

Next, we collect all valid 2D bounding boxes $\mathcal{B}^{2d}_{m,i}$ for each camera and their depth values $d^{m,i}_{centers}$ to a new set $\bm{\mathcal{V}}_m$.
A valid 2D bounding box $\mathcal{B}^{2d}_{m,i}$ should be partially or fully visible in the $m$-th camera and its depth  $d^{m,i}_{centers} \in [d_{min}, d_{max}]$.
We use $\bm{\mathcal{V}}_m$ to build up object-wise sparse depth maps. First, the raw depth map $\mathcal{D}'_m \in \mathbb{R}^{H_{depth} \times W_{depth}}$ of the $m$-th camera is initialized to zeros. Then for each pixel in $\mathcal{D}'_m$, we set the pixel value to the depth of the object center point if the pixel lies in an object's bounding box. If the pixel lies in multiple bounding boxes, we set it to the nearest one. 
Finally, the object-wise sparse depth map $\mathcal{D}_m$ is obtained by adopting linear-increasing discretization (LID) \cite{tang2020center3d} on $\mathcal{D}'_m$ for each camera. 

\subsection{Depth Predictor}
\label{depth predictor}
Inspired by previous depth-guided methods \cite{monodetr, monodtr} and depth estimation methods \cite{lss, fiery, bevdet, bevdet4d, caddn}, we utilize the Depth Predictor from MonoDETR \cite{monodetr} to learn depth information from object-wise sparse depth maps and extend this module to multi-view solution. To save the memory of our model, we use lightweight architecture built by convolution layers to predict depth distribution and match the number of depth bins as 3D positional embedding \cite{petr, petrv2}. Given image features $\mathcal{F}_{view}$, we use light-weight network $\mathbi{f}_{ddn}$ to predict depth logits $\mathcal{\hat{D}}$ and depth probability $\mathcal{\hat{D}}_{prob}$ among each depth map. Besides, we utilize Transformer encoder  $\psi_{i}$ to encode image features $\mathcal{F}_{view}$ into depth embeddings $\mathcal{F}_{depth}$ with multi-head attention $\psi_{i}$:
\begin{equation}
\begin{aligned}
    &\mathcal{E}_{0} = \mathcal{F}_{view},\\
    &\mathcal{E}_{i} = \psi_{i}(\mathcal{E}_{i-1}, \psi_{i-1}), i = 1,...,I,\\
    &\mathcal{F}_{depth} = \mathcal{E}_{I},
\end{aligned}
\end{equation}
where $\mathcal{E}_{i} \in \mathbb{R}^{\mathnormal{L} \times \mathnormal{C}}$ represents the depth embeddings from the Transformer encoder in the depth predictor. $\mathnormal{C}$ is the size of the embeddings. $\mathnormal{L} = \mathnormal{H}_{d} \times \mathnormal{W}_{d}$ is the length of depth embeddings. The number $i$ denotes the $i$-th layer in the encoder, and the encoder contains total $I$ layers. The final depth embedding is utilized in Sec. \ref{cross-depth attention}.

\subsection{Cross-view and Depth-guided Transformer}
\label{depth-guided transformer}
As the Transformer has successfully been used to fuse different modalities, we adopt it to combine both image features and depth embeddings. We adopt PETR \cite{petr} methods to conduct attention between different views. Besides, inspired by MonoDETR \cite{monodetr}, we insert depth embedding into multi-view attention from PETR \cite{petr}. To fuse different modalities, we extend the Transformer decoder layer to \textbf{Multi-attention Decoder Layer.}

\vspace{1em}
\noindent {\textbf{Multi-attention Decoder Layer.}}
We extend the Transformer decoder layer to the multi-attention decoder layer to conduct cross attention with both image features $\mathcal{F}_{view}$ and depth embeddings $\mathcal{F}_{depth}$. The cross-depth attention is passed first to provide candidate areas for object queries to propose objects.

\vspace{1em}
\noindent {\textbf{3D Object Query.}}
We modify the method from DETR3D \cite{detr3d} and PETR \cite{petr} and generate 3D object queries to decode bounding boxes and labels of candidates in 3D space. The 3D object queries are initialized uniformly in 3D space since the 3D reference points (in Sec. \ref{head and loss}) are sampled from 3D space. Those 3D object queries are unlike 2D DETR methods \cite{detr, deformabledetr, monodetr} and can be easier to learn for 3D Transformers (in Sec. \ref{depth-guided transformer}) since they provide 3D geometric hints.

\vspace{1em}
\noindent {\textbf{Cross-view Attention.}}
\label{cross-view attention}
We follow cross-view attention as methods from PETR \cite{petr}, and we feed image features $\mathcal{F}_{view}$ as keys and values. We utilize 3D positional embedding \cite{petr} as query positional embedding.

\vspace{1em}
\noindent {\textbf{Cross-depth Attention.}}
\label{cross-depth attention}
Previous methods \cite{petr, petrv2} only use visual information and thus lack depth cues for the detector. Inspired by MonoDETR \cite{monodetr},  MonoDTR \cite{monodtr} and other depth-guided methods \cite{astransformer, d4lcn, monogeo, monodtr, monodetr, deviant}, we suggest inserting depth hints as depth embedding into the detector to provide more detailed information for small objects. After we obtain depth embeddings $\mathcal{F}_{depth} \in \mathbb{R}^{\mathnormal{B} \times \mathnormal{N} \times \mathnormal{C} \times \mathnormal{H}_{d} \times \mathnormal{W}_{d} }$ from Sec. \ref{depth predictor}, we flatten depth embeddings into $\mathcal{F}_{depth} \in \mathbb{R}^{\mathnormal{\Acute{L}} \times \mathnormal{B} \times \mathnormal{C}}$ where $\mathnormal{\Acute{L}} = \mathnormal{N} \times \mathnormal{H}_{d} \times \mathnormal{W}_{d} $ indicates the flatten depth embeddings $\mathcal{F}_{depth}$ among all camera views. Then, we follow the previous Sec. \ref{cross-view attention} and select depth embeddings as keys and values for multi-head attention. Those depth embeddings can not only learn pixel-level depth hints in a single view in Depth Predictor (in Sec. \ref{depth predictor}) but also consider depth messages from the other views during cross attention mechanism.

\subsection{3D Detection Head and Loss}
\label{head and loss}
\noindent {\textbf{3D Detection Head.}}
To learn information between the camera view features and 3D position, we adopt the 3D Detection Head from PETR \cite{petr}. The 3D Detection Head from PETR \cite{petr} initialize 3D reference points among 3D space and adopt multi-layer perception to learn the candidate area in the ego-pose coordinate. Then, the 3D Detection Head generates 7 degrees of freedom to represent bounding boxes in the ego-pose coordinate.

\vspace{1em}
\noindent {\textbf{Detection Loss.}}
Following previous DETR-based methods \cite{detr3d, petr} for 3D object detection, we adopt Focal Loss \cite{focal} as classification loss $\mathbi{L}_{class}$ and $L1$ loss as bounding boxes regression loss $\mathbi{L}_{reg}$.

\vspace{1em}
\noindent {\textbf{Depth Distribution Network Loss.}}
To conduct the depth-guided method on a predefined depth map in Sec. \ref{object-wise sparse depth map}, we borrow the depth-guided method from MonoDETR \cite{monodetr} and refer to CaDDN \cite{caddn} and adopt Depth Distribution Network Loss (DDN Loss) to regularize the predicted depth map values and predicted depth map logits. Following CaDDN \cite{caddn}, we build our loss as the following equation:
\begin{equation}
    \mathbi{L}_{ddn} = \frac{1}{\mathnormal{W}_{d} \cdot \mathnormal{H}_{d}}\sum^{\mathnormal{W}_{d}}_{{u}=1}\sum^{\mathnormal{H}_{d}}_{{v}=1}\mathbi{FL}(\mathcal{D}({u},  {v}), \mathcal{\hat{D}}({u},  {v})),
\label{eq: loss_ddn}
\end{equation}
where $\mathnormal{W}_{d}$ and $\mathnormal{H}_{d}$ represent the size of depth map logits, and (${u}, {v}$) denotes the position of pixels. $\mathbi{FL}$ means adopted Focal Loss \cite{focal}.

Lastly, the total loss is computed by bipartite matching \cite{hungarian, deformabledetr, detr, detr3d, petr, petrv2} as below:
\begin{equation}
    \mathbi{L}_{total} = \mathcal{\alpha}_{class} \cdot \mathbi{L}_{class}+ \mathcal{\alpha}_{reg} \cdot \mathbi{L}_{reg}+ \mathcal{\alpha}_{ddn} \cdot \mathbi{L}_{ddn}.
    \label{eq: loss_total}
\end{equation}

\begin{figure*}[!t]
    \centering
    \includegraphics[width=1.0\textwidth, trim=0 0 0 0, clip]{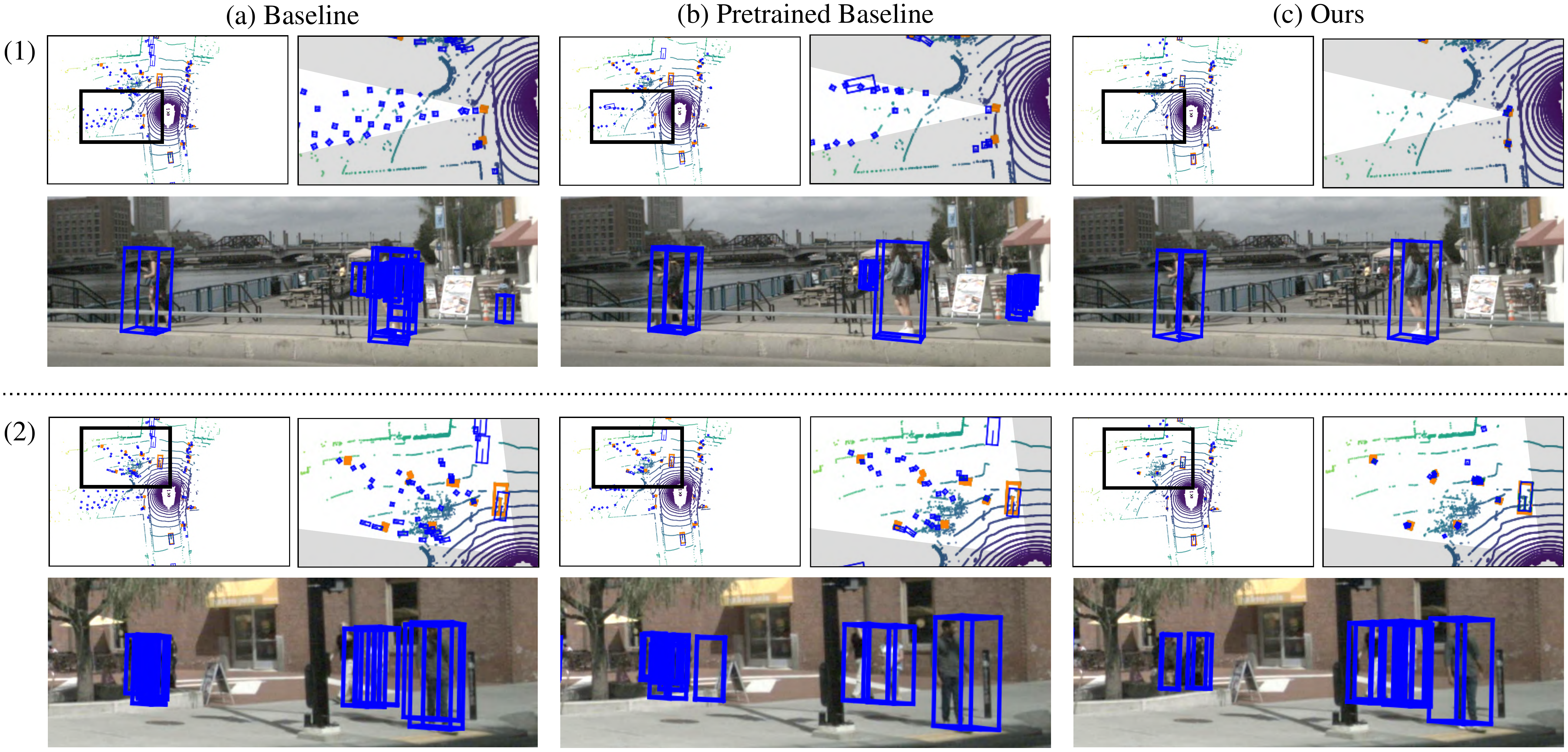}
    \caption{{\bf Visualization of false positive predictions on the nuScenes validation set.} We provide two qualitative examples, (1) and (2), with bird-eye-view (BEV) and camera-view representations. In the first row, the left and right images illustrate the focused areas of the global and zoomed-in BEVs, where blue and orange bounding boxes represent predictions and ground truth respectively. In the second row, the images in the camera view show predictions from models. Under a fair comparison with the same network backbone (PETR \cite{petr} with ResNet50 backbone \cite{resnet}), our cross-view and depth-guided method (c) effectively mitigates the false positive issue in prior multi-view baselines (a), i.e. {\bf does not produce repeated bounding boxes along the ray of depth}. Also, we even surpass methods equipped with a heavy-weight pretrained depth prediction module (b). Best viewed in color and zoom-in.}
    \label{fig: qualitative result}
\end{figure*}

\begin{table*}[!t]
    \centering
    \caption{{\bf Comparison with SOTA methods on the nuScene validation set.}{ \rm PETR\cite{petr} are trained with CBGS\cite{cbgs}. The best results are shown in {\bf bold}.}}
    \label{tab: sota all}
\resizebox{\linewidth}{!}{
\begin{tabular}{l|lll|ll|ll|lllll} 
\toprule
Methods   & Backbone  & Img Size & \#param.       & FPS↑         & GFLOPs↓        & mAP↑           & NDS↑           & mATE↓          & mASE↓          & mAOE↓          & mAVE↓          & mAAE↓          \\
\midrule
CenterNet\cite{centernet} & DLA       & 1600*900 & -              & -            & -              & 0.306          & 0.328          & {0.716} & 0.264          & 0.609          & 1.426          & 0.658          \\
FCOS3D\cite{fcos3d}    & ResNet101 & 1600*900 & {52.5M} & 1.7          & 2008.2         & 0.295          & 0.372          & 0.806          & 0.268          & 0.511          & 1.315          & \textbf{0.170} \\
PGD\cite{pgd}       & ResNet101 & 1600*900 & 53.6M          & 1.4          & 2223.0         & 0.335          & 0.409          & 0.732          & \textbf{0.263} & \textbf{0.423} & 1.285          & 0.172          \\
\midrule
Detr3D\cite{detr3d}    & ResNet101 & 1600*900 & \textbf{51.3M}          & 2.0          & 1016.8         & 0.303          & 0.374          & 0.860          & 0.278          & 0.437          & 0.967          & 0.235          \\
BEVDet\cite{bevdet}    & Swin-T    & 1408*512 & 126.6M         & 1.9          & 2962.6         & 0.349          & 0.417          & \textbf{0.637}          & 0.269          & 0.490          & 0.914          & 0.268          \\
PETR\cite{petr}      & ResNet101 & 1408*512 & 59.2M          & 5.3          & 504.6          & 0.357          & 0.421          & 0.710          & 0.270          & 0.490          & 0.885          & 0.224          \\
\midrule
CrossDTR      & ResNet101 & 1408*512 & 53.3M          & \textbf{5.8} & \textbf{483.9} & \textbf{0.370} & \textbf{0.426} & 0.773          & 0.269          & 0.482          & \textbf{0.866} & 0.203   
\\
\bottomrule
\end{tabular}
}
\end{table*}

\section{Experiments}

\begin{table}[t]
    \caption{{\bf Comparison with lightweight multi-view methods. }{\rm We utilize {\bf bold} to highlight the best results.}}
    \label{tab: sota multi-cam}
    \centering
\resizebox{\linewidth}{!}{
\begin{tabular}{l|ll|ll|l}
\toprule
Methods   & Backbone  & \#param.       & FPS↑          & GFLOPs↓        & mAP↑           \\
\midrule
DETR3D\cite{detr3d}    & ResNet101 & 51.3M          & 2.0           & 1016.8         & 0.303          \\
BEVDet\cite{bevdet}    & ResNet50  & 54.1M          & 9.3           & 452.0          & 0.299          \\
BEVFormer\cite{bevformer} & ResNet50  & 68.7M          & 2.3           & 1303.5         & 0.252          \\
PETR\cite{petr}      & ResNet50  & 36.6M          & 10.4          & 297.2          & 0.317          \\
\midrule
CrossDTR      & ResNet50  & \textbf{31.8M} & \textbf{10.6} & \textbf{268.1} & \textbf{0.326}\\
\bottomrule
\end{tabular}
}
\end{table}
\begin{table}[!t]
    \centering
    \caption{{\bf Ablation study of depth-guided module}. {\rm DE denotes Depth Embedding and DDN Loss denotes Depth Distribution Network Loss. Our method was built on PETR\cite{petr} with DDN Loss and DE.}}
    \label{tab: ablations}
\begin{tabular}{l|ll|ll}
\toprule

Methods & DE                    & DDN Loss              & mAP↑           & NDS↑           \\
\midrule
PETR\cite{petr}    &                       &                       & 0.357          & 0.421          \\
\midrule
CrossDTR  & \multicolumn{1}{c}{$\checkmark$} &                       & 0.366          & 0.423          \\
CrossDTR  & \multicolumn{1}{c}{$\checkmark$} & \multicolumn{1}{c|}{$\checkmark$} & \textbf{0.370} & \textbf{0.426} \\

\bottomrule
\end{tabular}
\end{table}
\begin{table}[t]
    \caption{{\bf Study of false positive predictions for the pedestrian class. }{\rm We choose PETR\cite{petr} with ResNet50 \cite{resnet} and PETR\cite{petr} with depth-pretrained VoVNetV2 \cite{vovnetv2} as baselines and compare them with our method, CrossDTR, with different distance thresholds. We utilize {\bf bold} to highlight the best results.}}
    \label{tab: false positive}
    \centering
\resizebox{\linewidth}{!}{
\begin{tabular}{l|ll|lll}
\toprule
\multirow{2}{*}{Methods} & \multirow{2}{*}{Depth-pretrained} & \multirow{2}{*}{Backbone} & \multicolumn{3}{l}{AP (Pedestrain) @ Dist.↑}               \\ \cmidrule{4-6}

                        &                                   &                           & [0.5]      & [1.0]      & [4.0]     \\
\midrule
\multirow{2}{*}{PETR\cite{petr}}   &                                   & ResNet50                  & 0.09           & 0.401          & 0.809          \\
                        & \multicolumn{1}{c}{$\checkmark$}                                 & VoVNetV2                  & 0.102          & 0.426          & 0.870          \\
\midrule
CrossDTR                    &                                   & ResNet50                  & \textbf{0.320} & \textbf{0.689} & \textbf{0.875}\\
\bottomrule
\end{tabular}
}
\end{table}

\subsection{Setup}
\noindent {\textbf{Dataset.}}
\label{sec: dataset}
In this paper, we use the nuScenes dataset \cite{nuscenes} as our benchmark, which provides camera, radar, and LiDAR sensor data with 3D bounding box annotations. Its data is mainly composed of camera data and provides only sparse LiDAR data as an auxiliary. As it lacks data to provide depth information like dense LiDAR or depth maps, we generate object-wise sparse depth maps (in Sec. \ref{object-wise sparse depth map} for the cross-view and depth-guided method. The nuScenes dataset \cite{nuscenes} contains 1000 scenes and each scene is 20 seconds in length and annotated at 2HZ.

\vspace{1em}
\noindent {\textbf{Implementation Details.}}
\label{sec: implementation details}
Following training policies from \cite{detr3d, petr, petrv2}, we use features from backbones \cite{resnet, swintransformer, vovnet, vovnetv2} with downsample scale of $\frac{1}{16}$ and $\frac{1}{32}$. And we feed the features into both depth predictor and cross-view attention. Besides, we follow \cite{petr, petrv2} to sample 64 points for depth in 3D positional embeddings and also for depth predictor to estimate depth distribution. Specifically, we supervise generated depth maps only during training. We use the AdamW optimizer with a learning rate of 2e-4 to train our model. We train our model for 24 epochs on 4 Nvidia 3090 GPUS with a total batch size of 8 for 48 hours. We use input images at resolution $512 \times 1408$ for our baseline model.

\vspace{1em}
\noindent {\textbf{Baselines.}}
\label{sec: baselines}
We compare CrossDTR with both monocular and multi-camera approaches. CenterNet\cite{centernet}, FCOS3D \cite{fcos3d}, and PGD \cite{pgd} represent monocular approaches, while DETR3D \cite{detr3d}, PETR \cite{petr}, and BEVDet \cite{bevdet} serve as the baselines of multi-camera ones. Please note that, for fair comparison, we only adopt the performance of these approaches without tricks such as test-time augmentation \cite{fcos3d, pgd}, CBGS \cite{cbgs}, and oversampling \cite{cbgs}. Besides, the comparison with the lightweight BEVFormer \cite{bevformer} (from their official repository) without encoding extra temporal information is also included.

\vspace{0.5em}
\noindent {\textbf{Evaluation Metrics.}}
\label{sec: evaluation metrics}
We report the seven true positive metrics, mean Average Translation Error (mATE), mean Average Scale Error (mASE), mean Average Orientation Error (mAOE), mean Average Velocity Error (mAVE), mean Average Attribute Error (mAAE), mean Average Precision (mAP), and nuScenes detection score (NDS), which is the aggregation of the above metrics, defined in the nuScenes dataset \cite{nuscenes}. mAP estimates the distance between the centers of a predicted and a ground-truth 3D bounding box. To evaluate the efficacy of our method in solving false positive predictions, we report the Average Precision of \emph{pedestrian} as our metrics. We also take Frame Per Second (FPS) and Giga Flops (GFLOPs) into consideration to evaluate the real-time ability of multi-camera models.

\subsection{Quantitative Results}
\label{sec: quantitative results}
\noindent {\textbf{Comparison with State-of-the-art.}}
As shown in Tab. \ref{tab: sota all}, our method surpasses other previous methods and achieves the state-of-the-art performance of mAP and NDS on the validation dataset \cite{nuscenes}. To begin with, CenterNet \cite{centernet}, FCOS3D \cite{fcos3d}, and PGD\cite{pgd} are classic monocular baseline. Our method exceeds by more than \textbf{3 percent on mAP and 2 percent on NDS}. Additionally, compared with the SOTA multi-camera methods (starting from the fifth row), our method still surpasses all of them by at least \textbf{1.3 percent on mAP and 0.5 percent on NDS}. Swin-T represents Swin-Transformer \cite{swintransformer}, which is the strongest backbone among the Tab. \ref{tab: sota all}. Our method with ResNet101 also beats BEVDet \cite{bevdet} with Swin-T \cite{swintransformer}. Then, our method needs the least computational resource \textbf{(483.9 GFLOPs and 5.8 FPS)}. Our method is lightweight and can conduct real-time 3D detection on the nuScenes \cite{nuscenes} dataset.

\vspace{1em}
\noindent {\textbf{Comparison with lightweight multi-view methods.}}
Tab. \ref{tab: sota multi-cam} shows the comparison between our method and previous multi-camera methods. Note that all the scores are from their official repositories. Since we conduct our experiments on the validation set with limited computation resources, we choose a smaller backbone ResNet50 \cite{resnet} to extract features from input images at resolution $512 \times 1408$. Our proposed method overtakes all previous multi-camera methods, even against DETR3D \cite{detr3d} with stronger ResNet101 backbone \cite{resnet} and BEVFormer \cite{bevformer} with temporal information. Our method surpasses the second best method by \textbf{0.9 percent on mAP and 1.1 percent on NDS}. Besides, our model contains the least parameters as shown in Tab. \ref{tab: sota multi-cam} and attains the highest score \textbf{(10.6) on FPS}. The result shows that our model can be made into practice to conduct real-time detection.

\subsection{Ablation Study}
\label{sec: ablation study}
Tab. \ref{tab: ablations} shows the effectiveness of our cross-view and depth-guided module. We conduct an ablation study to verify the effectiveness of depth embedding (DE) and Depth Distribution Network Loss (DDN Loss) on the validation set. We compare different architectures with the baseline model PETR \cite{petr}. We find the performance is improved by {0.9 percent on mAP and 0.2 percent on NDS} when we plug depth embeddings into the cross-attention, and the full model achieves the best performance increasing by \textbf{1.3 percent on mAP and 0.5 percent on NDS}.

\subsection{False Positive Predictions Results}
\label{sec: false positive}
\begin{figure}[!t]
    \centering
    \includegraphics[width=0.4\textwidth, trim=0 0 0 0, clip]{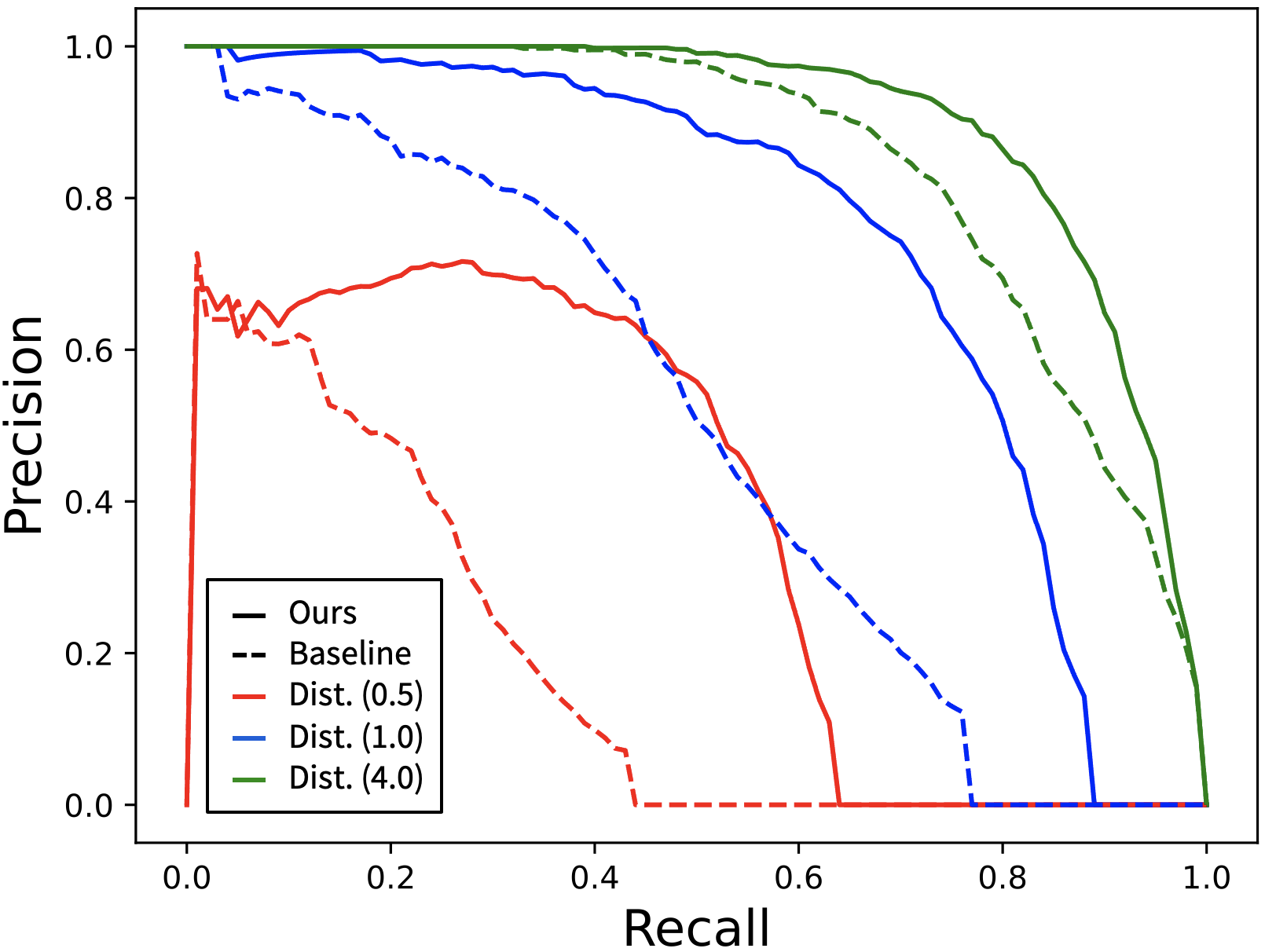}
    \caption{{\bf The precision-recall curve of pedestrian class.} Fig. \ref{fig: pr_curve_pedestrian} shows the comparison of AP between baseline (dotted lines) and our method (solid lines). The red, blue, and green colors represent the distance threshold at 0.5, 1.0, and 4.0 respectively. Regardless of distance, our method hugely outperforms the baseline on small object (e.g. pedestrian) detection.}
    \label{fig: pr_curve_pedestrian}
\end{figure}

To verify whether our method can resolve the false positive problem, we consider Average Precision (AP). Tab. \ref{tab: false positive} shows the AP of the pedestrian class with different distance thresholds on the validation set. Our method surpasses our baseline with a depth-pretrained backbone by \textbf{over 10 percent on average} among each threshold. Moreover, Fig. \ref{fig: pr_curve_pedestrian} illustrates our overall performance predominantly exceeds the baseline on all thresholds and thus resolves the false positive issue. Red, blue, and green represent the distance threshold at 0.5, 1.0, and 4.0 respectively. Note that the initial vertical red line is caused by unstable training and overlapping bounding boxes.

\subsection{Qualitative Results}
\label{sec: qualitative results}
Fig. \ref{fig: qualitative result} shows the qualitative result. Orange and blue bounding boxes represent ground truth and predictions respectively. As shown in Fig. \ref{fig: qualitative result}, both PETR \cite{petr} with ResNet50 backbone \cite{resnet} and PETR \cite{petr} with depth-pretrained VoVNetV2 \cite{vovnet, vovnetv2, ddad, dd3d} still predict a row of false positive predictions along the direction of depth for small objects. Since depth-pretrained backbones are generally pretrained on the external dataset and contain different settings on camera matrices, we suggest that those backbones can narrowly deal with the false positive problem due to weak depth estimation. Nevertheless, our method can predominately alleviate this problem due to referred depth information from the internal dataset \cite{nuscenes}.

\section{Conclusion}
In this paper, we design an end-to-end Cross-view and Depth-guided Transformer, called CrossDTR, for 3D object detection. To address the false positive bounding boxes commonly existing in prior multi-view approaches, a lightweight Depth Predictor, supervised by our produced object-wise sparse depth maps, is proposed to generate low-dimensional depth embeddings. Furthermore, to combine image and depth hints from different views, a Cross-view and Depth-guided Transformer is developed to fuse this information efficiently. We are optimistic that our proposed method would pave a new way for developing a cost-effective and real-time 3D object detector.
 
\section{Acknowledgement}
This work was supported in part by the National Science and Technology Council, under Grant MOST 110-2634-F-002-051, Qualcomm through a Taiwan University Research Collaboration Project, and Mobile Drive Technology Co., Ltd (MobileDrive). We are grateful to the National Center for High-performance Computing.



\clearpage

\bibliographystyle{IEEEtran}
\bibliography{reference}

\end{document}